\documentclass[letterpaper, 10 pt, conference]{ieeeconf}  % Comment this line out if you need a4paper

\IEEEoverridecommandlockouts                              % This command is only needed if 
\title{\LARGE \bf
\textsc{ActPrune}: Action-Grounded Visual Token Pruning for Efficient Vision-Language-Action Models
}

\author{Sungwon Moon$^{1}$, Minji Kim$^{2}$, Hyemin Boo$^{2}$, Daeun Song$^{2,\dagger}$, and Jiyoung Lee$^{1,2,\dagger}$%
\thanks{$^{1}$Human-Centered Artificial Intelligence Research Institute, Ewha Womans University  
        {\tt \{sungwon268,lee.jiyoung\} @ewha.ac.kr}}%
\thanks{$^{2}$Division of Artificial Intelligence \& Software, Ewha Womans University  
        {\tt \{kim.minji03,hyeminb,songd,lee.jiyoung\} @ewha.ac.kr}}%
\thanks{$^{\dagger}$Corresponding authors}%
}
\usepackage{tabularx}
\usepackage{amsmath}
\usepackage{amssymb}  % assumes amsmath package installed
\usepackage{makecell}
\usepackage{array} 
\usepackage{multirow}
\usepackage{lipsum}
\usepackage{booktabs}
\usepackage{color}
\usepackage{colortbl}
\usepackage{spverbatim}
\usepackage{listings}
\usepackage[hidelinks]{hyperref}
\usepackage[capitalize]{cleveref}
\usepackage{arydshln}
\usepackage{subfig}
\usepackage[percent]{overpic}

\definecolor{codekeyword}{rgb}{0.0, 0.0, 0.5}   % Dark blue for keywords
\definecolor{codecomment}{rgb}{0.0, 0.5, 0.0}   % Dark green for comments
\definecolor{codestring}{rgb}{0.56, 0.0, 1.0}   % Purple for strings
\definecolor{backcolour}{rgb}{0.98,0.98,0.97}

\definecolor{lastcol}{gray}{0.95} 

\lstdefinestyle{pythonstyle}{
    language=Python,                          % Set language to Python
    basicstyle=\ttfamily\small,               % Basic font style
    keywordstyle=\color{codekeyword}\bfseries,% Keywords in bold dark blue
    commentstyle=\color{codecomment}\itshape, % Comments in italic dark green
    stringstyle=\color{codestring},           % Strings in purple
    showstringspaces=false,                   % Do not show spaces in strings
    breaklines=true,                          % Automatic line breaking
    tabsize=4,                                % Set tab size
    numbers=none,                             % No line numbers
    frame=single, % No frame around code
    backgroundcolor=\color{backcolour},
    captionpos=b,                             % Caption position at bottom
    morekeywords={self, __init__, __name__, __main__}, % Additional keywords
}

\definecolor{codegray}{gray}{0.95}

\definecolor{baselinecolor}{gray}{.9}

\newcommand{\ul}[1]{\underline{#1}}

\newcommand{\paragrapht}[1]{\noindent\textbf{#1}}  % tidy \paragraph

\usepackage{pifont}
\usepackage{amssymb}

\definecolor{turquoise}{cmyk}{0.65,0,0.1,0.3}
\definecolor{purple}{rgb}{0.65,0,0.65}
\definecolor{green}{rgb}{0, 0.5, 0}
\definecolor{orange}{rgb}{0.8, 0.6, 0.2}
\definecolor{red}{rgb}{0.8, 0.2, 0.2}
\definecolor{darkred}{rgb}{0.6, 0.1, 0.05}
\definecolor{blueish}{rgb}{0.0, 0.3, .6}
\definecolor{light_gray}{rgb}{0.7, 0.7, .7}
\definecolor{pink}{rgb}{1, 0, 1}
\definecolor{greyblue}{rgb}{0.25, 0.25, 1}

\newcommand{\cmark}{\color{blueish}{\ding{51}}}%
\newcommand{\xmark}{\color{red}{\ding{55}}}%

\usepackage{xspace}
\usepackage{array}
\usepackage{layouts}
\usepackage{algorithm}
\usepackage{bbding}
\usepackage{listings}
\usepackage{pifont}
\usepackage{wasysym}
\usepackage{cite}

\usepackage{footnote}
\makesavenoteenv{tabular}
\makesavenoteenv{table}
\makesavenoteenv{figure}
\usepackage{pythonhighlight}

\newcommand{\ours}{\textsc{ActPrune}\xspace}
\newcommand{\projector}{\mathcal{F}_{\text{proj}}}
\newcommand{\venc}{\mathcal{F}_{\text{enc}}}
\newcommand{\mG}{\mathcal{G}}

\newcommand{\pend}{\mathbf{p}^{\text{ee}}}
\newcommand{\tildepend}{\tilde{\mathbf{p}}^{\text{ee}}}
\newcommand{\pendcam}{\mathbf{p}^{\text{w}}}

\newcommand{\Dt}{\Delta\mathbf{t}}
\newcommand{\DR}{\Delta\mathbf{R}}

\newcommand{\bt}{\mathbf{t}}

\newcommand{\bR}{\mathbf{R}}
\newcommand{\bK}{\mathbf{K}}

\newcommand{\bg}{\mathbf{g}}
\newcommand{\bv}{\mathbf{v}}
\newcommand{\bV}{\mathbf{V}}
\newcommand{\bX}{\mathbf{X}}
\newcommand{\bA}{\mathbf{A}}

\newcommand{\ba}{\mathbf{a}}
\newcommand{\bz}{\mathbf{z}}
\newcommand{\ie}{\textit{i.e.\xspace}}
\newcommand{\eg}{\textit{e.g.\xspace}}

\usepackage{xcolor}
\definecolor{mygreen}{RGB}{65,159,102}
\definecolor{myred}{RGB}{200,86,88}
\definecolor{dark_green}{rgb}{0, 0.5, 0}

\usepackage[most]{tcolorbox}
\newtcolorbox{promptbox}{colback=gray!5, colframe=gray!40, boxrule=0.4pt, arc=1pt, left=4pt, right=4pt, top=3pt, bottom=3pt, fontupper=\small\ttfamily}

\begin{document}

\maketitle
\thispagestyle{empty}
\pagestyle{empty}

\begin{abstract}
Vision-language-action (VLA) models perform versatile manipulation tasks from language instructions, but their inference cost limits how fast a robot can respond in closed-loop control.
Visual token pruning reduces this cost; however, prior methods estimate token redundancy from encoded visual embeddings or attention, only after the visual encoder has processed the entire observation.
We propose \ours, a training-free token pruning method that moves the pruning decision ahead of the vision encoder by grounding it in a geometric prior derived from the robot's own action.
The action chunk predicted at the previous control step is already available before the current observation is encoded, and it specifies the end-effector and camera motion between consecutive observations.
The same action affects the two standard camera views asymmetrically: in the fixed third-person view, \ours retains patches along the projected end-effector trajectory together with semantic anchors, while in the wrist view, it prunes patches that remain unchanged after compensating for the action-induced camera motion.
Extensive experiments on two simulation environments (LIBERO and VLABench) and four real-robot tasks demonstrate that \ours achieves up to 1.61 $\times$ speedup and 59.5\% FLOPs reduction with negligible degradation in success rates. The project site is \url{http://mmai.ewha.ac.kr/actprune/}.
\end{abstract}
\begin{figure}[ht]
    \centering
    \subfloat[]{
    \includegraphics[width=.9\linewidth]{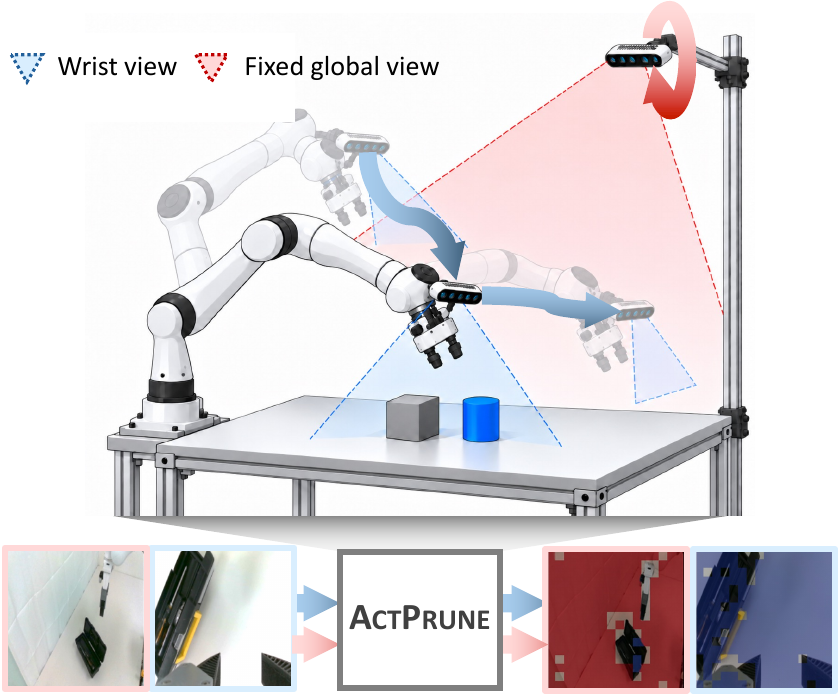}
    }\\ \vspace{-1em}
    \subfloat[]{
    \includegraphics[width=.8\linewidth]{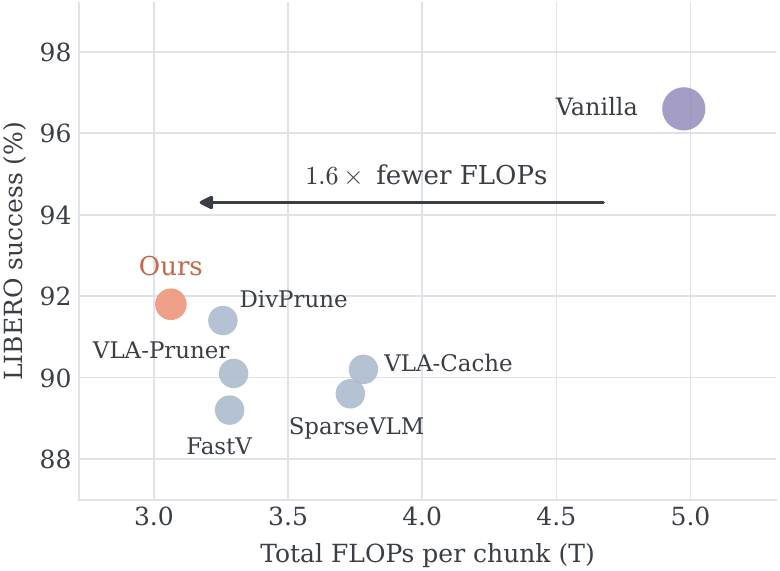}
    }
    \caption{
    Overview of \ours. (a) By grounding token pruning in executed actions, (b) \ours achieves the largest FLOPs reduction and highest speedup at 25\% token retention in a training-free manner, while preserving competitive task performance.
    }
    \label{fig:placeholder}
    \vspace{-1em}
\end{figure}

\section{Introduction}

A long-standing goal in robotics is to develop a single policy that can perform diverse manipulation tasks from natural language instructions.
Vision-language-action (VLA) models have made progress toward this goal. Many of them build on pretrained vision-language models and generalize to new objects and instructions by inheriting web-scale representations. However, this capability comes with high inference cost. VLA models rely on large vision-language backbones that process hundreds of visual tokens at every control step. In closed-loop control, this computation is repeated for every new observation. As a result, inference latency can limit how fast the robot can respond.

% visual token pruning using visual information 
Recent methods have explored different ways to reduce the inference cost of VLA models~\cite{robomamba, tinyvla, qail, deervla}. Visual token pruning instead reduces computation by removing redundant visual tokens during inference. 
Existing pruning methods identify redundant tokens from encoded visual features~\cite{divprune, visionzip}, attention~\cite{fastv, sparsevlm}, temporal changes~\cite{vlacache, specprune}, or action information~\cite{vlapruner, specprune}.
In these approaches, redundancy is mainly determined from the visual observations themselves, after the vision encoder has processed every token. 

% VLA models inherit the architecture of their multimodal large language model (MLLM) backbones, where visual observations are encoded into a sequence of tokens that are subsequently processed with language to predict actions. The number of visual tokens therefore affects the computational cost of each control step.
% Recent works~\cite{fastv, } have reduced that cost by pruning or merging redundant tokens. 
% Most existing methods determine token importance from encoded visual features~\cite{divprune,visionzip} or language-model attention~\cite{sparsevlm}. These signals determine visual redundancy primarily from the observation itself.
%operate on tokens that the vision encoder has already produced.
%We propose \ours to revisit VLA models~\cite{?} as embodied systems.

% action connects consecutive observations
A robot in closed-loop control, however, does not simply observe a sequence of images. Between two observations, the robot executes an action that changes its state and interaction with the scene. The action therefore provides information about where and how the next observation may change. More importantly, this action is already available before the next observation arrives. Unlike visual features or attention scores, it does not require the current observation to be encoded first. The previous action can therefore provide an early signal for identifying visual redundancy before visual encoding.%This allows us to reason about visual changes between observations using the robot's own action. 
% Rather than using the action only as the output of the policy, we use it to understand how the next observation changes. 

% A policy in a control loop, however, is not a passive observer of an image stream.
% Its own action induces predictable changes between consecutive observations. 
% Moreover, that action is available before the next observation reaches the encoder. 
% Existing VLA token reduction methods do not exploit this action-induced structure. 
% This suggests a different way of using the policy's action for visual token reduction. 
% Prior work reads the action backward, scoring tokens by how strongly the action head attended to them~\cite{vlapruner}.
% But the action might also be read forward.
% The policy in the current step determines the camera pose and the scene configuration at the next step.
% Therefore, the policy indicates which regions of the next observation carry new information, before that observation is encoded.

% The second property concerns the asymmetry between camera views.
% How this action-induced change appears in the observation depends on the camera view.
The effect of the action on the observation depends on the camera viewpoint. 
A standard manipulation setup pairs a fixed third-person camera with a wrist-mounted camera~\cite{droid, libero, vlabench}.
In the fixed third-person view, the scene stays stable across steps.
Change is confined to the region the manipulator occupies.
In contrast, the wrist view changes globally and dynamically across steps. 
These changes arise from camera motion and viewpoint shifts induced by the actions.
The action therefore implies contrasting patterns of new information across camera views.

% We propose \ours, a training-free framework that reads the policy's own actions forward to prune visual tokens before encoding.
Based on these insights, we propose \ours (Fig.~\ref{fig:placeholder}), a training-free framework that exploits the closed-loop structure of robot manipulation to reduce visual computation. 
Specifically, the action chunk predicted at the previous step is available before the current observation is processed in the visual encoder.
It provides a physically grounded prior on where visual computation should be spent without requiring a vision encoder forward pass.
\ours leverages this prior to prune visual tokens, projecting the anticipated end-effector trajectory onto the fixed third-person view to mark newly visualized information, while removing motion-compensated redundant patches for the wrist view.
The remaining budget is applied after the encoder, where encoder attention ranks the retained tokens.
% The key idea is to turn action-induced observation changes into a view-specific pruning signal.
% These changes mark new information in the fixed view, but in the wrist view they explain what the policy has already seen.
% Two mechanisms complement this action-derived signal.
% Pruned tokens are spatially merged rather than discarded to preserve scene context.
% We further defer part of the pruning budget to a post-encoder stage, where encoder attention provides an additional measure of visual importance.

We summarize our contributions as follows: 
\begin{enumerate}
    \item We introduce \ours, a training-free framework that derives geometric priors directly from robot actions, enabling visual token pruning before the vision encoder.
    \item We identify that the same action induces distinct redundancy patterns in fixed and wrist-mounted views, and exploit this asymmetry with view-specific pruning.
    \item \ours achieves state-of-the-art efficiency on LIBERO~\cite{libero} and VLABench~\cite{vlabench}, with up to 1.61$\times$ speedup and 59.5\% FLOPs reduction while preserving competitive performance. We further validate its real-world advantage on a 7-DoF Franka Research 3.
    
\end{enumerate}

 \section{Related Work}

\subsection{Efficient Vision-Language-Action Models}\vspace{-3pt}
VLA models enable robots to perform general-purpose manipulation tasks. 
However, they require substantial computation at every control step, and inference latency limits the control frequency.
Existing approaches improve VLA efficiency at different levels of the inference pipeline. 
Lightweight architectures~\cite{robomamba, tinyvla} replace or compress components of the backbone, quantization methods~\cite{qail} reduce the numerical precision of model parameters, and early-exit methods~\cite{deervla} terminate computation once
intermediate representations suffice.
While effective, these methods require retraining or modification of the underlying model. 

Another line of work targets action generation rather than reducing the cost of individual policy inference. 
High-frequency VLA frameworks~\cite{openvlaoft, pi0fast} increase control frequency through action chunking or asynchronous decoding.
By predicting multiple actions at once or decoupling execution from inference, these methods reduce how frequently the full policy must be invoked. 
However, each policy inference still processes the vision-language backbone, making per-inference visual computation an important bottleneck. 
Our work is complementary to these approaches. Rather than modifying the policy architecture or action-generation process, we reduce the visual computation required during closed-loop inference without additional training. 
\vspace{-3pt}

\subsection{Visual Token Pruning for VLA}\vspace{-3pt}
Visual token pruning has been widely explored in VLMs to reduce this cost.
Existing methods estimate token importance using encoded visual features~\cite{prumerge, visionzip, divprune} or language-model attention~\cite{fastv, sparsevlm}. Tokens with low estimated importance are then pruned or merged before subsequent transformer layers. 
This reduces language-model computation by limiting the number of visual tokens.
These approaches are effective for static vision-language inputs, but typically determine redundancy from each observation independently. 
% However, each observation is processed independently.
They therefore do not exploit the temporal structure of closed-loop robot control.

Recent VLA acceleration methods extend visual token pruning to closed-loop control.
Temporal redundancy can be identified from changes across consecutive observations~\cite{vlacache, specprune}.
Frame differencing provides a lightweight measure of this redundancy, but becomes unreliable under camera motion because it compares pixels at fixed image coordinates.
Action information has also been incorporated into token selection~\cite{specprune, vlaadp, vlapruner}.
In these methods, actions are combined with signals obtained after visual encoding to estimate token importance.
As a result, token selection occurs only after visual encoding.
In contrast, \ours uses the executed action to identify redundancy in the subsequent observation before visual encoding.

\begin{figure*}
    \centering
    \includegraphics[width=.9\linewidth]{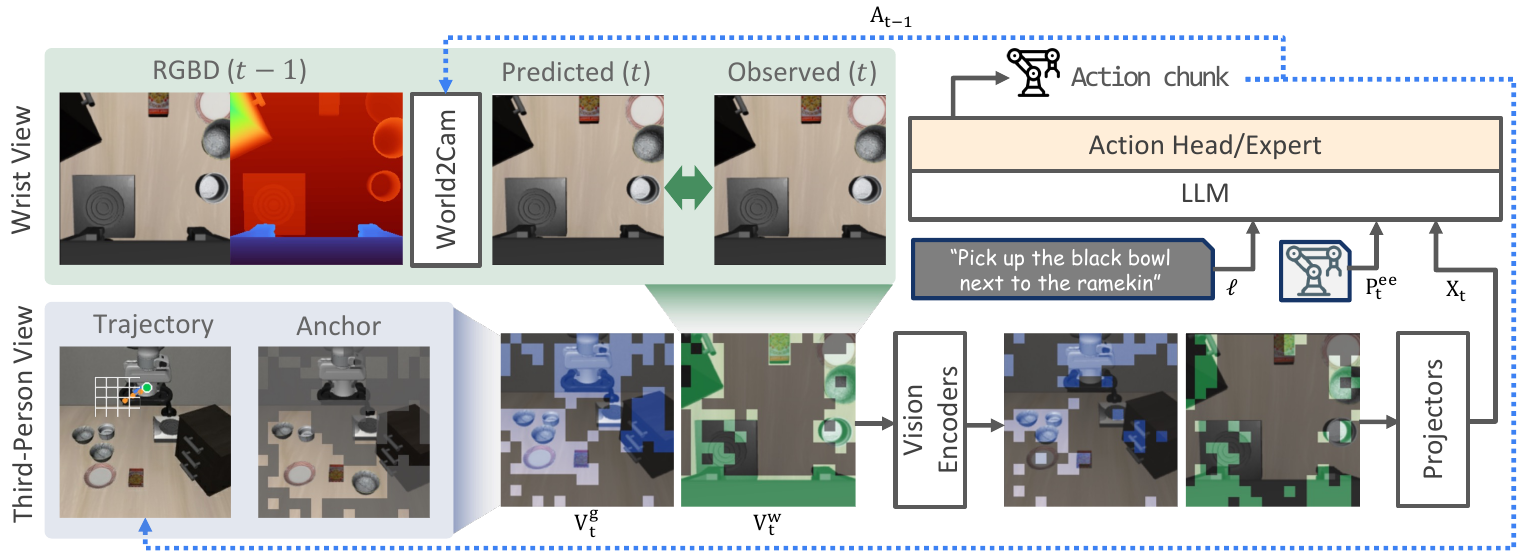}
    \vspace{-.7em}
    \caption{The overall framework of \ours. The previous action chunk $\bA_{t-1}$ derives view-specific pruning signals before visual encoding. Pruned patches are compressed into context tokens, and encoder attention applies the remaining budget.}
    \label{fig:alg}
\end{figure*}

\section{Preliminaries}
VLA models~\cite{openvla,openvlaoft, pi05} build upon MLLMs~\cite{llama2, gemma2} and are trained on large-scale robot demonstrations, generating executable robot actions from visual observations and task instructions.
At each time step $t$, the observation consists of a fixed third-person view and a wrist view, along with a task instruction $\ell$ (\eg, ``\texttt{Pick a towel and place it in the basket}'').
Each view is divided into a grid of non-overlapping patches, where each patch embedding $\bV_t^g = \{\bv_{t,k}^g\}_{k=1}^{N_v}$ for the fixed third-person view and $\bV_t^w = \{\bv_{t,k}^w\}_{k=1}^{N_v}$ for the wrist view.
Vision encoders $\venc$ (\ie, SigLIP~\cite{siglip} or DINOv2~\cite{dino}) followed by projectors $\projector$ map the patches of each view into visual tokens $\bX_t \in \mathbb{R}^{2N_v \times D}$:
\begin{equation}
\bX_t
=
\projector
\left(
\venc(\bV_t^g) \oplus \venc(\bV_t^w)
\right),
\end{equation}
where $\oplus$ denotes the concatenation operation.
% where $N$ is the 
% where $X_t^{\mathrm{vis}}\in\mathbb{R}^{N \times D}$ and $X_t^{\mathrm{txt}}\in\mathbb{R}^{L_{\mathrm{txt}}\times D}$ denote the visual and language embeddings, respectively. 
% We denote the end-effector pose $\pend$.
% , and its embedding by $X_t^{\mathrm{prop}}$.
The VLA policy $\mG$ predicts an action chunk $\bA_t=\{\ba_t^i, \bg_t^i\}_{i=1}^{N_{\text{act}}}$ from the action placeholder tokens or noise $\bA_{\emptyset}$,
\begin{equation}
    \mathbf{A}_t= \mG(\bX_t, \ell, \bA_{\emptyset}, \pend_t),
\end{equation}
where $\pend_t$ denotes the end-effector pose in world coordinate, and each action $\ba$ is represented as
\begin{equation}
\ba_t^i
=
[\,\DR,\,\Dt]^\top
\in \mathbb{R}^{6},
\label{eq:action_dim}
\end{equation}
where $\DR\in\mathbb{R}^{3}$ and $\Dt\in\mathbb{R}^{3}$ denote the rotational and translational components, respectively, and $\bg_t^i \in \{0,1\}$ denotes the gripper state.
% During closed-loop control, only a prefix of $A_t$ is executed before the next observation is acquired.
% We denote this executed prefix from the previous timestep by $\bar{A}_{t-1}$.
% Thus, $\bar{A}_{t-1}$ is available before the next visual observations $I_t^s$ and $I_t^w$ are processed by the vision encoder.

\section{Method}

\subsection{Overview \& Problem Formulation}
As illustrated in \cref{fig:alg}, \ours leverages the previous action $\bA_{t-1}$ to derive view-specific pruning signals before visual encoding.
The same action affects the two views asymmetrically: (i) in the third-person view, it reveals where the manipulator will move within a static scene.
\ours projects the end-effector trajectory from $\bA_{t-1}$ onto the image plane and retains patches around it.
Semantic anchors, extracted once from the first observation, additionally preserve task-relevant regions.
(ii) In the wrist view, the action instead moves the camera itself, so temporal redundancy under motion compensation reveals prunable regions.
Previous methods~\cite{vlacache,specprune} compare patches at corresponding spatial locations across consecutive frames, where even static regions exhibit large appearance changes under the dynamically translated wrist view.
In contrast, \ours compensates for the camera motion induced by $\bA_{t-1}$ by reprojecting patches from the previous observation into the current view using depth maps from the wrist-mounted RGBD camera.
Patches whose features remain unchanged after compensation carry no additional information beyond the previous observation.
\ours therefore merges them into context tokens that preserve their coarse content.
% We read $\bar{A}_{t-1}$ forward and leverage the asymmetry between camera views to derive view-specific pruning signals before visual encoding.
% The actions indicate where new information may appear in the scene-fixed view and what has already been observed in the end-effector-mounted view.

 \begin{table*}[ht]
\centering
\small
\caption{
{Performance comparison with OpenVLA-OFT on LIBERO} at 50\%, 25\%, and 12.5\% visual token retention. \textbf{The best} is bold, and \ul{the second} is underlined.
}
\label{tab:main_results_libero_openvla}
\begin{tabular}{l cccc cccc}
\toprule
Method & Spatial & Object & Goal & Long & Avg.(\%)$\uparrow$ & {FLOPs(T)$\downarrow$} & {Latency(ms)$\downarrow$} & {Speedup($\times$)$\uparrow$}\\
\midrule

% \rowcolor{gray!15}
% \multicolumn{9}{l}{\textit{Retained 100\% (oracle)}} \\

OpenVLA-OFT (100\%)
& 99.0 & 97.0 & 97.4 & 92.6 & 96.5 & 4.739 & 86.173 & 1.00$\times$ \\

% \rowcolor{gray!15}
\midrule
\multicolumn{9}{l}{\textit{Retained 50\% (50\%$\downarrow$)}} \\

\hspace{1em}+FastV
& \textbf{98.6} & 97.6 & \ul{97.2} & 91.2 & 96.2 & 3.123 & \ul{66.19} & \ul{1.30}$\times$ \\

\hspace{1em}+SparseVLM 
& \ul{98.0} & 96.2 & \ul{97.2} & 80.8 & 93.1 & 3.388 & 71.89 & 1.20$\times$ \\

\hspace{1em}+DivPrune
& 97.8 & \ul{98.2} & \ul{97.2} & \ul{92.8} & \ul{96.5} & \ul{3.015} & 72.33 & 1.19$\times$ \\

\hspace{1em}+VLA-Cache
& 97.0 & 96.2 & 97.1 & 90.6 & 95.2 & 3.509 & 78.72 & 1.09$\times$ \\

\hspace{1em}+VLA-Pruner
& 97.6 & 97.4 & 96.4 & 92.6 & 96.0 & 3.025 & 73.49 & 1.17$\times$ \\

\hspace{1em}+SpecPrune-VLA 
& 95.2 & \textbf{98.4} & \textbf{97.4} & 48.0 & 84.8 & 3.069 & 72.05 & 1.20$\times$ \\

\hspace{1em}+\textbf{\ours}
& \textbf{98.6} & \textbf{98.4} & \ul{97.2} & \textbf{93.4} & \textbf{96.9} & \textbf{2.814} & \textbf{65.04} & \textbf{1.33}$\times$ \\

\midrule
\multicolumn{9}{l}{\textit{Retained 25\% (75\%$\downarrow$)}} \\

\hspace{1em}+FastV 
& \ul{98.4} & 97.0 & 96.4 & \ul{89.8} & 95.4 & 2.325 & \ul{54.62} & \ul{1.58}$\times$ \\

\hspace{1em}+SparseVLM 
& 97.8 & 97.2 & \textbf{97.6} & 74.4 & 91.8 & 2.666 & 61.74 & 1.40$\times$ \\

\hspace{1em}+DivPrune 
& \textbf{98.6} & 97.8 & 95.4 & 88.8 & 95.2 & \ul{2.172} & 55.23 & 1.56$\times$ \\

\hspace{1em}+VLA-Cache 
& 98.0 & 86.0 & 95.4 & 79.2 & 89.7 & 2.869 & 63.20 & 1.36$\times$ \\

\hspace{1em}+VLA-Pruner 
& 98.0 & 97.6 & 97.0 & \ul{89.8} & \ul{95.6} & 2.240 & 63.11 & 1.37$\times$ \\

\hspace{1em}+SpecPrune-VLA 
& 94.4 & \ul{98.4} & 94.8 & 44.0 & 82.9 & 2.283 & 54.37 & 1.58$\times$ \\

\hspace{1em}+\textbf{\ours}
& 98.0 & \textbf{98.8} & \ul{97.4} & \textbf{90.0} & \textbf{96.1} & \textbf{1.921} & \textbf{53.61} & \textbf{1.61}$\times$ \\

\midrule
\multicolumn{9}{l}{\textit{Retained 12.5\% (87.5\%$\downarrow$)}} \\

\hspace{1em}+FastV 
& \textbf{98.0} & 90.8 & \textbf{94.4} & 64.2 & 86.9 & 1.941 & 52.91 & 1.63$\times$ \\

\hspace{1em}+SparseVLM 
& \ul{97.2} & 94.6 & 92.6 & 65.6 & 87.5 & 2.340 & 59.94 & 1.44$\times$ \\

\hspace{1em}+DivPrune 
& 93.6 & 90.0 & 85.6 & 53.0 & 80.6 & \ul{1.750} & \ul{52.51} & \ul{1.64}$\times$ \\

\hspace{1em}+VLA-Cache
& 92.2 & 88.2 & 92.4 & 56.4 & 82.3 & 2.517 & 59.08 & 1.46$\times$ \\

\hspace{1em}+VLA-Pruner 
& 96.8 & 95.2 & 91.8 & \ul{68.8} & \ul{88.2} & 1.844 & 65.28 & 1.32$\times$ \\

\hspace{1em}+SpecPrune-VLA
& 94.0 & \textbf{97.4} & \ul{94.0} & 40.8 & 81.6 & 2.018 & 54.35 & 1.59$\times$ \\

\hspace{1em}+\textbf{\ours}
& 96.0 & \ul{96.0} & 91.0 & \textbf{71.2} & \textbf{88.6} & \textbf{1.474} & \textbf{52.02} & \textbf{1.66}$\times$ \\

\bottomrule
\end{tabular}
\vspace{-.2em}
\end{table*}
 
Specifically, given a target retained ratio $\gamma$, we allocate the retained token budget into a patch-level ratio $\gamma_{\text{patch}}$ and an embedding-level ratio $\gamma_{\text{embed}}$, such that $\gamma = \gamma_{\text{patch}} + \gamma_{\text{embed}}$.
Consequently, only $2N_v\cdot \gamma$ visual tokens are forwarded to the LLM backbone; since visual tokens constitute the vast majority of the input sequence compared to the short instruction (\ie, $N_v >> N_\ell$), this directly translates into a substantial reduction in VLA inference cost.

\subsection{Patch-level Pruning in the Third-person View}
The static third-person view captures the overall spatial layout of the workspace, including the manipulator, the target objects, and their relations.
Since the camera is static, the informative changes in this view are induced solely by the robot's own motion.
The previously predicted chunk $\bA_{t-1}$ specifies where such changes will occur:
\begin{equation}
    \pend_{(t-1) \rightarrow j}
    = \pend_{t-1} + \sum_{i=1}^{j} \Dt_{t-1}^{i},
    \qquad j = 1, \ldots, N_{\text{act}},
\label{eq:traj}
\end{equation}
We project the position $\pend_{(t-1) \rightarrow j}$ of each anticipated pose onto the image plane using the camera intrinsic and extrinsic parameters:
\begin{equation}
    \bar{\bv}^{\text{g}}_{\,j} = \pi\!\left(\bK\,[\bR_{ \text{w} \rightarrow \text{c}} \mid \bt_{\text{w} \rightarrow \text{c}}]\, \tildepend_{(t-1) \rightarrow j}\right),
\label{eq:traj_proj}
\end{equation}
where $[\bR_{ \text{w} \rightarrow \text{c}} \mid \bt_{\text{w} \rightarrow \text{c}}]$ is the world-to-camera rotation and translation, $\tildepend$ is the homogeneous coordinate of $\pend$, $\bK$ is the camera intrinsics matrix, and $\pi([x,y,z]^\top) = [x/z,\, y/z]^\top$ is the perspective projection.
Among these poses, we use the $n_p$ most recently executed ones and extend the trajectory by $n_f$ poses extrapolated along the recent motion direction. 
Patches within the $\kappa\times\kappa$ neighborhoods of the projected positions are retained to preserve the manipulator's workspace and nearby interaction regions.
When the latest gripper command indicates a closed gripper (\ie, $\bg_{t-1}$=$1$), we enlarge $\kappa$ to account for the manipulated object, as ablated in \cref{tab:gripper}).

In addition, we build semantic anchors from the first observation using the decoder's attention, and reuse their spatial locations throughout the whole episode.
It therefore preserves task-relevant regions that remain static.
We carefully notice that such anchors are computed once per episode and introduce no additional per-step encoding.
Trajectory regions are prioritized, with semantic anchors filling the remaining budget.

\subsection{Patch-level Pruning in the Wrist View}
The camera is rigidly attached to the end-effector, with a fixed hand-eye transformation relating the camera frame to the end-effector frame.
Starting from the previous camera pose $\pendcam_{t-1}$, we integrate the translation and rotation $\ba_{t-1}^i$ to obtain the relative camera transformation:
\begin{equation}
    \pendcam_{(t-1) \rightarrow t} = \pendcam_{t-1} + \ba_{t-1}^i.
\end{equation}
For each patch $\bv_k$ in the previous observation with depth $\bz_k$, we back-project it into 3D, transform it by $\pendcam_{(t-1) \rightarrow t} = [\bR' \mid \bt']$, and project it onto the current image plane:
\begin{equation}
    \bar{\bv}^{\text{w}}_k = \pi\!\left(\bK \left(\bR' \,\bz_k \bK^{-1}\tilde{\bv}^{\text{w}}_k
    + \bt' \right)\right),
\end{equation}
where $\tilde{\bv}$ is the homogeneous coordinate of $\bv$.
Note that patches with projections outside the image are retained.

For each correspondence $(\bv^{\text{w}}_k, \bar{\bv}^{\text{w}}_k)$ established by the reprojection, we compute the cosine similarity to retain $\gamma_{\text{patch}}$ of patches with the lowest score.
Namely, temporally redundant patches (\eg, background) are pruned to avoid the computational overhead of visual encoding.
Rather than being discarded entirely, the pruned patches are merged into coarse context tokens: the image is partitioned into an $M\times M$ grid, and the pruned patches within each cell are averaged into a single token (ablated in \cref{tab:grid}).
Since tokens are compressed through lightweight geometric operations before visual encoding, \ours reduces the computation of both the vision encoder and the LLM, unlike prior methods that prune only after encoding.
% A high $s_i$ indicates temporal redundancy after compensating for camera motion.
% We rank the patches by $s_i$ and remove the most redundant patches until the pre-encoder pruning budget $\gamma_{\text{pre}}$ is reached.

% \paragraph{Preserving Context.}

% Spatial context is important for identifying objects and their relations in manipulation.
% Rather than discarding pruned patches, we compress them into coarse context tokens.
% Specifically, we divide the image grid into $M\times M$ cells and average the patch embeddings selected for pruning within each cell.
% Each context token is assigned the position of the patch nearest to the cell center.
% These tokens are retained within the vision encoder but discarded before the language model.

\subsection{Embedding-level Pruning}
While patch-level pruning exploits action-induced changes across observations, the attention weights obtained by vision encoders instead capture the relative importance of patches within the current observation~\cite{vlacache,vlapruner, specprune}.
We therefore defer part of the pruning budget to the post-encoder stage, where the retained tokens are ranked by encoder attention and pruned using the remaining budget $\gamma_{\text{embed}}$.
\begin{figure}[t]
    \centering
    \includegraphics[width=\linewidth]{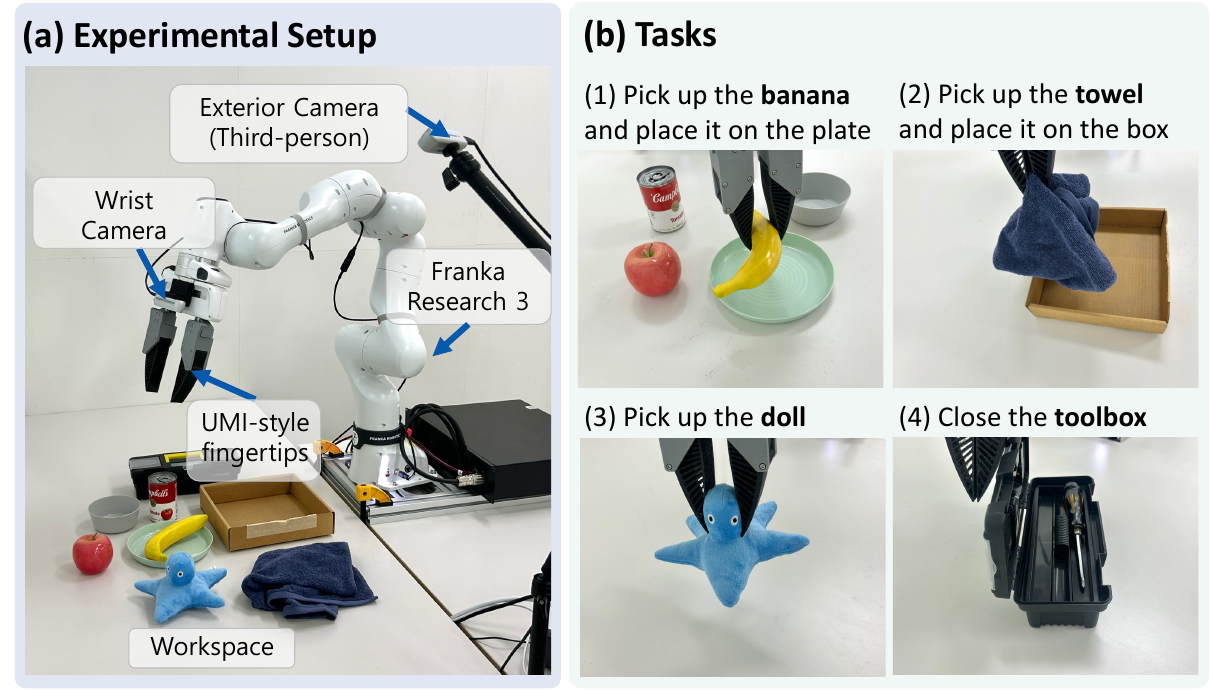}\vspace{-0.3em}
    \caption{Real-world experimental setup. (a) A 7-DoF Franka Research 3 with UMI-style fingertips, observed by an exterior third-person camera and a wrist-mounted camera. (b) Four manipulation tasks.}
    \vspace{-1em}
    \label{fig:real_tasks}
\end{figure}
\section{Experiments}

We evaluate \ours on simulation and real-world manipulation tasks across VLA models. We compare against token pruning methods and ablate key design choices.

 \begin{table*}[t]
\centering
\small
\caption{
{Performance comparison with $\pi_{0.5}$ on LIBERO} at 50\%, 25\%, and 12.5\% visual token retention. 
 % \textbf{The best} is bold, and \ul{the second} is underlined.
}
\label{tab:main_results_libero_pi}
\begin{tabular}{l cccc cccc}
\toprule
% \multirow{2}[3]{*}{Method}
% & \multirow{2}[3]{*}{Avg.(\%)$\uparrow$}
% & \multirow{2}[3]{*}{FLOPs(T)$\downarrow$}
% & \multirow{2}[3]{*}{Latency(ms)$\downarrow$}
% & \multirow{2}[3]{*}{Speedup($\times$)$\uparrow$} \\
% \cmidrule(lr){2-5}
Method & Spatial & Object & Goal & Long & Avg.(\%)$\uparrow$ & {FLOPs(T)$\downarrow$} & {Latency(ms)$\downarrow$} & {Speedup($\times$)$\uparrow$}\\
\midrule
% \rowcolor{gray!15}
% \multicolumn{9}{l}{\textit{Retained 100\% (oracle)}} \\
$\pi_{0.5}$ (100\%)
& 98.0 & 98.0 & 99.4 & 90.8 & 96.6 & 4.975 & 178.3 & 1.00$\times$ \\
% \rowcolor{gray!15}
\midrule
\multicolumn{9}{l}{\textit{Retained 50\% (50\%$\downarrow$)}} \\
\hspace{1em}+FastV
& 97.4 & \textbf{98.6} & 96.2 & 89.6 & 95.5 & 3.952 & 178.1 & 1.00$\times$ \\
\hspace{1em}+SparseVLM
& \textbf{98.4} & 98.2 & 96.2 & \textbf{91.2} & \underline{96.0} & 4.143 & \underline{177.1} & \underline{1.01$\times$} \\
\hspace{1em}+DivPrune
& \underline{98.2} & 96.6 & 96.8 & 88.8 & 95.1 & \underline{3.825} & 180.1 & 0.99$\times$ \\
\hspace{1em}+VLA-Cache
& 98.0 & 97.0 & 95.2 & 90.2 & 95.1 & 4.091 & 181.4 & 0.98$\times$ \\
\hspace{1em}+VLA-Pruner
& 98.0 & 97.8 & \underline{97.0} & 89.0 & 95.5 & 3.895 & 180.3 & 0.99$\times$ \\
\hspace{1em}+\textbf{\ours}
& 98.0 & \underline{98.4} & \textbf{98.4} & \underline{90.6} & \textbf{96.4} & \textbf{3.695} & \textbf{175.5} & \textbf{1.02$\times$} \\

\midrule
\multicolumn{9}{l}{\textit{Retained 25\% (75\%$\downarrow$)}} \\
\hspace{1em}+FastV
& 88.4 & \underline{97.8} & 87.0 & 83.6 & 89.2 & 3.447 & \underline{173.8} & \textbf{1.03$\times$} \\
\hspace{1em}+SparseVLM
& 89.6 & \underline{97.8} & 86.4 & \underline{84.6} & 89.6 & 3.632 & 175.0 & 1.02$\times$ \\
\hspace{1em}+DivPrune
& \textbf{93.6} & 95.0 & 91.2 & \textbf{85.8} & \underline{91.4} & 3.257 & 175.8 & 1.01$\times$ \\
\hspace{1em}+VLA-Cache
& 91.0 & 94.8 & \textbf{92.0} & 83.0 & 90.2 & 3.781 & 177.4 & 1.01$\times$ \\
\hspace{1em}+VLA-Pruner
& 90.4 & 97.6 & 88.8 & 83.4 & 90.1 & \underline{3.357} & 176.6 & 1.01$\times$ \\
\hspace{1em}+\textbf{\ours}
& \underline{92.0} & \textbf{99.0} & \underline{91.4} & \underline{84.6} & \textbf{91.8} & \textbf{3.063} & \textbf{172.9} & \textbf{1.03$\times$} \\

\midrule
\multicolumn{9}{l}{\textit{Retained 12.5\% (87.5\%$\downarrow$)}} \\
\hspace{1em}+FastV
& 68.4 & \underline{89.6} & 64.0 & 50.4 & 68.1 & 3.196 & \underline{172.4} & \underline{1.03$\times$} \\
\hspace{1em}+SparseVLM
& 67.6 & 85.8 & 63.2 & 54.2 & 67.7 & 3.357 & 173.3 & 1.03$\times$ \\
\hspace{1em}+DivPrune
& \underline{72.2} & 87.2 & 70.2 & \underline{58.4} & \underline{72.0} & \underline{2.975} & 175.2 & 1.02$\times$ \\
\hspace{1em}+VLA-Cache
& 71.4 & 85.8 & 69.6 & 51.4 & 69.6 & 3.627 & 176.3 & 1.01$\times$ \\
\hspace{1em}+VLA-Pruner
& 70.2 & 86.8 & \textbf{73.2} & 55.2 & 71.4 & 3.075 & 175.9 & 1.01$\times$ \\
\hspace{1em}+\textbf{\ours}
& \textbf{74.0} & \textbf{90.2} & \underline{72.4} & \textbf{59.4} & \textbf{74.0} & \textbf{2.734} & \textbf{171.7} & \textbf{1.04$\times$} \\
\bottomrule
\end{tabular}
\vspace{-.5em}
\end{table*}
\begin{figure*}[h]
    \centering
    \subfloat[Spatial: Pick up the black bowl between the plate and the ramekin and place it on the plate]{
        \centering
        \includegraphics[width=.48\linewidth]{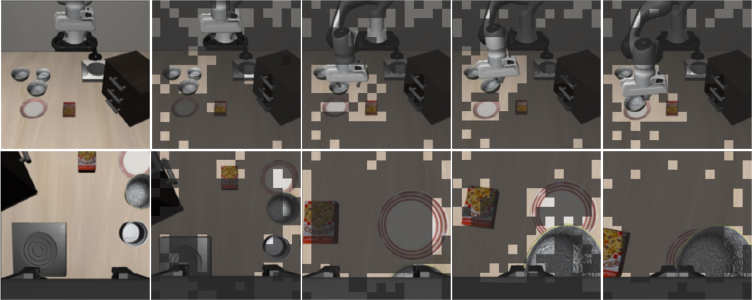}
    }
    \subfloat[Object: Pick up the alphabet soup and place it in the basket]{
        \centering
        \includegraphics[width=.48\linewidth]{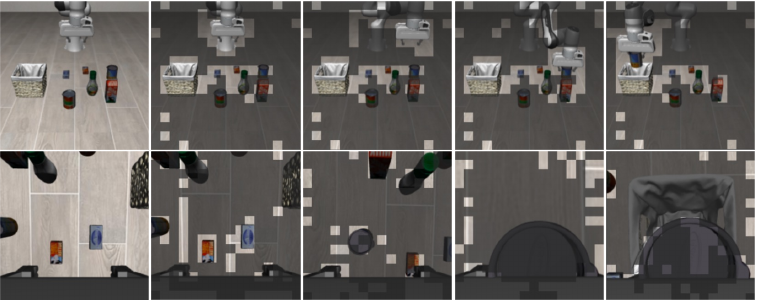}
    }\\ 
    \vspace{-.7em}
    \subfloat[Goal: {Open the middle drawer of the cabinet}]{
        \centering
        \includegraphics[width=0.48\linewidth]{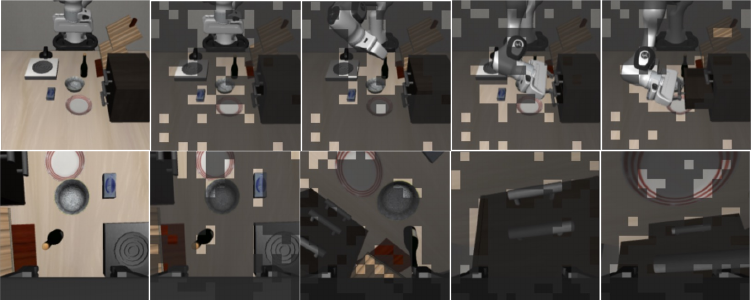}
    }
    \subfloat[Long: {Put both the alphabet soup and the tomato sauce in the basket}]{
        \centering
        \includegraphics[width=0.48\linewidth]{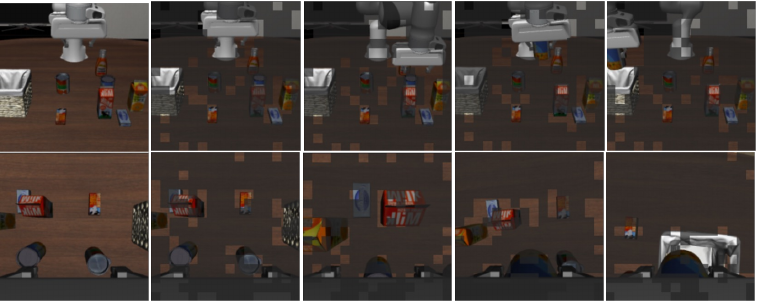}
    }\\
    \vspace{-.3em}
    \caption{Qualitative results of retained (color) and merged/pruned (gray) tokens on the four LIBERO tasks at 25\% retained ratio.
    The leftmost column shows the observation, and the subsequent columns visualize the pruning results over the episode, with the third-person view on the top row and the wrist view on the bottom row.}
    \label{fig:libero}
    \vspace{-1em}
\end{figure*}

\subsection{Experimental Setup}\vspace{-3pt}

% We evaluate \ours on both simulation benchmarks and real-world manipulation tasks. We compare against existing visual token reduction methods and conduct ablation studies to evaluate its effectiveness and analyze its design choices. 

\paragrapht{Baselines.}
We evaluate \ours with two VLA models, OpenVLA-OFT~\cite{openvlaoft} and $\pi_{0.5}$~\cite{pi05}, comparing against both general-purpose visual token pruning and VLA-specific acceleration methods.
General-purpose baselines are FastV~\cite{fastv}, SparseVLM~\cite{sparsevlm}, and DivPrune~\cite{divprune}, which prune by attention or feature statistics.
VLA-specific baselines are VLA-Cache~\cite{vlacache}, VLA-Pruner~\cite{vlapruner}, and SpecPrune-VLA~\cite{specprune}, which additionally use temporal or action information.

\paragrapht{Evaluation Metrics.}
We evaluate all methods according to retained ratios using three metrics: success rate, FLOPs, and latency speedup. Success rate measures task performance under closed-loop control, while FLOPs measure the computational cost of the full VLA model, including the vision encoder and language-model backbone. We report speedup over the vanilla model based on end-to-end inference latency, including the overhead introduced by token pruning.

\begin{table*}[t]
\centering
\small
\caption{
{Performance comparison with $\pi_{0.5}$ on VLABench} at 50\%, 25\%, and 12.5\% visual token retention. Task names are abbreviated for brevity (e.g., ‘Con.’ for ‘add condiment’, ‘Chem.’ for ‘select chemistry tube’).}
\label{tab:vlabench_main}
% \resizebox{\textwidth}{!}{
\setlength{\tabcolsep}{3pt}
\begin{tabular}{l cccccccccc cc}
\toprule
{Method}
& {Con.} & {Flower} & {Book} & {Chem.}
& {Drink} & {Fruit} & {Mahj.} & {Paint.}
& {Poker} & {Toy} & {Avg.(\%)} & {FLOPs(T)$\downarrow$} \\
\midrule

% \multicolumn{12}{l}{\textit{Retained 100\% (oracle)}} \\
$\pi_{0.5}$ (100\%)
& 60.0 & 20.0 & 51.0 & 57.8 & 33.3 & 40.0 & 60.0 & 20.0 & 60.0 & 32.0& 42.0 & 4.975 \\

\midrule
\multicolumn{12}{l}{\textit{Retained 50\% (50\%$\downarrow$)}} \\

\hspace{1em}+FastV
& 56.0 & 6.0 & 27.1 & 36.2 & 28.0 & 30.0 & 8.5 & 23.8 & 34.0 & 6.0 & 25.6 & 3.447 \\

\hspace{1em}+SparseVLM
& 28.0 & 2.0 & 12.5 & 18.6 & 22.2 & 25.0 & 9.1 & 33.3 & 6.0 & 5.0 & 16.2 & 3.542 \\

\hspace{1em}+DivPrune
& 52.0 & 6.0 & 24.8 & 32.8 & 26.8 & 40.0 & 23.4 & 28.6 & 30.0 & 16.0 & 28.0 & 3.257  \\

\hspace{1em}+VLA-Cache
& 55.0 & 0.0 & 34.4 & 30.0 & 22.2 & 30.0 & 33.3 & 20.0 & 30.0 & 10.0 & 26.5 & 3.674 \\

\hspace{1em}+VLA-Pruner
& 50.0 & 20.0 & 33.3 & 35.5 & 20.0 & 50.0 & 40.0 & 25.6 & 20.0 & 0.0 & 29.4 & 3.357 \\

\hspace{1em}+\textbf{\ours}
& 62.0 & 6.0 & 34.0 & 23.4 & 26.5 & 26.0 & 22.4 & 28.5 & 30.0 & 4.0 & 26.3 & 3.019  \\

\midrule
\multicolumn{12}{l}{\textit{Retained 25\% (75\%$\downarrow$)}} \\

\hspace{1em}+FastV
& 48.0 & 0.0 & 25.0 & 22.9 & 26.0 & 2.0 & 2.0 & 22.8 & 0.0 & 0.0 & 14.9 & 2.698 \\

\hspace{1em}+SparseVLM
& 14.0 & 0.0 & 4.3 & 0.0 & 24.4 & 2.0 & 2.0 & 19.5 & 0.0 & 0.0 & 6.6 & 2.896  \\

\hspace{1em}+DivPrune
& 38.0 & 4.0 & 17.9 & 12.0 & 14.0 & 28.0 & 12.9 & 25.5 & 12.0 & 4.0 & 16.8 & 2.415  \\

\hspace{1em}+VLA-Cache
& 20.0 & 0.0 & 20.0 & 20.0 & 22.2 & 10.0 & 17.5 & 10.0 & 20.0 & 10.0 & 14.9 & 2.515 \\

\hspace{1em}+VLA-Pruner
& 50.0 & 10.0 & 22.2 & 0.0 & 10.0 & 0.0 & 11.1 & 23.8 & 0.0 & 0.0  & 12.7 & 2.815 \\

\hspace{1em}+\textbf{\ours}
& 30.0 & 0.0 & 25.5 & 6.5 & 14.2 & 14.0 & 8.3 & 38.0 & 0.0 & 0.0 & 13.7 & 2.063 \\

\bottomrule
\end{tabular}
\vspace{-.5em}
\end{table*}

\paragrapht{Real Robot Setup.} 
Our setup consists of a 7-DoF Franka Research 3 robot equipped with a Franka gripper with UMI~\cite{polyumi,umi}-style fingertips, and two Intel RealSense D435 cameras providing third-person and wrist views. Policy inference is performed on NVIDIA RTX PRO 4500 Blackwell. Following $\pi_{0.5}$, we initialize the policy from the DROID-pretrained checkpoint~\cite{droid} and fine-tune it on our demonstrations using LoRA for 20K steps.

\vspace{-3pt}
\subsection{Evaluation Benchmarks}\vspace{-3pt}

\paragrapht{LIBERO}~\cite{libero} consists of four task suites: Spatial, Object, Goal, and Long. Each suite contains 10 tasks, and we evaluate 50 episodes per task, resulting in 500 episodes per suite. We follow the evaluation protocol of OpenVLA-OFT~\cite{openvlaoft} and use the official model weights.

\paragrapht{VLABench}~\cite{vlabench} is a large-scale language-conditioned manipulation benchmark comprising 100 task categories and over 2,000 objects with extensive domain randomization. We evaluate 10 primitive tasks with 50 episodes per task using the official $\pi_{0.5}$ checkpoint fine-tuned on these tasks.

\paragrapht{Real-world Benchmark.}
We further evaluate the real-world performance of \ours on four manipulation tasks consisting of placing a \textbf{banana} on a plate, picking up a \textbf{doll}, closing a \textbf{tool}box, and placing a \textbf{towel} on a box, as shown in \cref{fig:real_tasks}.
These tasks span different object properties and interaction types: rigid, deformable, and articulated objects, and both pick-and-place and in-place manipulation.
For each task, we conduct 10 trials with variations in object poses and initial configurations. Task success is determined according to predefined task-specific completion criteria.
\vspace{-3pt}
%Our setup consists of a 7-DoF Franka Research 3 robot equipped with a Franka gripper, an Intel RealSense D435 third-person camera, and an Intel RealSense D435 wrist-mounted camera. Following the fine-tuning protocol of $\pi_{0.5}$, we initialize the policy from the DROID-pretrained checkpoint~\cite{droid} and fine-tune it on our demonstrations using LoRA for ?k steps.

\begin{table}[!t]
\centering
\small
\caption{
Analysis of design choices for \ours on LIBERO with $\pi_{0.5}$ are reported. The retained ratio is 25\%.
}
\label{tab:ablation_choice}
\setlength{\tabcolsep}{3pt}
\begin{tabular}{cc cc cc}
\toprule
\multicolumn{2}{c}{Third-person View}
& \multicolumn{2}{c}{Wrist View}
& \multirow{2}[3]{*}{Acc.(\%)$\uparrow$}
& \multirow{2}[3]{*}{FLOPs(T)$\downarrow$} \\
\cmidrule(lr){1-2} \cmidrule(lr){3-4}
Patch & Emb. & Patch & Emb. & & \\
\midrule
\cmark & \xmark & \xmark & \xmark & 93.5 & 3.990 \\
\cmark & \cmark & \xmark & \xmark & 94.8 & 4.044 \\
\xmark & \xmark & \cmark & \xmark & 94.9 & 4.100 \\
\xmark & \xmark & \cmark & \cmark & 96.0 & 4.058 \\
\cmark & \cmark & \cmark & \cmark & 91.8 & 3.063 \\
\bottomrule

\end{tabular}%
\vspace{-.7em}
\end{table}

\subsection{Main Results}\vspace{-3pt}
% table 1~3
To evaluate whether \ours preserves task performance under visual token reduction, we compare it with existing token reduction methods across different VLA models and manipulation benchmarks. 
In particular, \cref{tab:main_results_libero_openvla} and \cref{tab:main_results_libero_pi} report the results on LIBERO with OpenVLA-OFT and $\pi_{0.5}$, respectively. On OpenVLA-OFT, \ours consistently achieves the highest average success rate across all retention ratios, reaching 96.9\%, 96.1\%, and 88.6\% at 50\%, 25\%, and 12.5\% retention, respectively. 
The advantage is particularly evident at 25\% retention, where \ours retains performance close to the vanilla policy while reducing FLOPs from 4.739 to 1.921 and achieving a 1.61$\times$ speedup. At 12.5\% retention, \ours further reduces FLOPs to 1.474 while maintaining the highest average success rate among the compared methods. 
The same trend holds for $\pi_{0.5}$. As shown in \cref{tab:main_results_libero_pi}, \ours achieves the highest average success rate at all three retention ratios. 
\cref{fig:libero} visualizes the retained tokens (colored patches) in \ours. Even at a 25\% retained ratio, they concentrate on the workspace and the target objects (\eg,  \texttt{alphabet soup}, \texttt{basket}).

We further assess \ours on VLABench using $\pi_{0.5}$, as reported in \cref{tab:vlabench_main}. At 50\% retention, \ours achieves an average success rate of 26.3\%, which remains competitive with existing pruning methods. At 25\% retention, performance degrades substantially for all methods, with \ours remaining competitive with the VLA-specific baselines. 
Across both settings, \ours attains the lowest FLOPs, reducing them to 2.063 at 25\% retention.
Overall, the results demonstrate that \ours effectively reduces visual computation while preserving manipulation performance across different VLA models and benchmarks.

\begin{table*}[t]
\centering
\caption{\ours ablation experiments with $\pi_{0.5}$ on LIBERO at retained ratio 25\%.}
\label{tab:ablations}\vspace{-.5em}

\subfloat[$(n_p, n_f)$ denotes the number of past and extrapolated future trajectory points used for patch-level pruning in the third-person view. Balanced trajectory achieves the best.
% \vspace{\baselineskip}
\label{tab:horizon}]{
\centering
\begin{minipage}{0.48\textwidth}{\begin{center}
\begin{tabular}{c cccc c}
\toprule
\multirow{2}[3]{*}{$(n_p, n_f)$}
& \multicolumn{4}{c}{LIBERO Success (\%)}
& \multirow{2}{*}{Acc.(\%) $\uparrow$} \\
\cmidrule(lr){2-5}
& Spatial & Object & Goal & Long & \\
\midrule
(5, 0) & \textbf{92.8} & 97.6 & 89.4 & \textbf{86.0} & 91.5 \\

\rowcolor{gray!15}
(3, 3) & 92.0 & \textbf{99.0} & \textbf{91.4} & 84.6 & \textbf{91.8} \\

(0, 5) & 92.2 & 98.0 & 88.6 & 82.0 & 90.2 \\
\bottomrule
\end{tabular}%
\end{center}}
\end{minipage}
}
\hspace{2em}
\subfloat[
$(\kappa_{\mathrm{o}}, \kappa_{\mathrm{c}})$ are the sizes used when the gripper is open and closed, respectively. When the gripper is closed, the larger $\kappa_{\mathrm{c}}$ covers the grasped object carried along the trajectory.
\label{tab:gripper}
]{
\begin{minipage}{0.45\textwidth}{\begin{center}
\begin{tabular}{c cccc c}
\toprule
\multirow{2}[3]{*}{$(\kappa_{\mathrm{o}},\kappa_{\mathrm{c}})$}
& \multicolumn{4}{c}{LIBERO Success (\%)}
& \multirow{2}{*}{Acc.(\%)$\uparrow$} \\
\cmidrule(lr){2-5}
& Spatial & Object & Goal & Long & \\
\midrule
(3, 5) & 90.4 & 98.0 & 87.8 & 85.4 & 90.4 \\
(5, 5) & 91.6 & 98.8 & 85.8 & 83.0 & 89.8 \\
\rowcolor{gray!15}
(5, 7) & 92.0 & \textbf{99.0} & \textbf{91.4} & 84.6 & \textbf{91.8} \\
(7, 9) & \textbf{93.2} & 97.8 & 88.0 & \textbf{85.6} & 91.2 \\
\bottomrule
\end{tabular}%
\end{center}}
\end{minipage}
}
\vspace{-2em}
\end{table*}
\begin{table*}[t]
\centering
\caption{\ours ablation experiments with OpenVLA-OFT on LIBERO at retained ratio 25\%.}
\label{tab:ablations_oft}\vspace{-.5em}

\subfloat[Token merging on visual encoders. Especially, DINOv2 drops performance more when pruning than merging.
% \vspace{\baselineskip}
\label{tab:merging}]{
\centering
\begin{minipage}{0.52\textwidth}{\begin{center}
% \vspace{-1em}
\begin{tabular}{cc cccc c}
\toprule
\multicolumn{2}{c}{Merging}
& \multicolumn{4}{c}{LIBERO Success (\%)}
& \multirow{2}{*}{Acc.(\%)$\uparrow$} \\
\cmidrule(lr){1-2} \cmidrule(lr){3-6}
DINOv2 & SigLIP & Spatial & Object & Goal & Long & \\
\midrule				
\xmark & \xmark & 98.2 & 96.8 & 96.6 & 88.2 & 95.0 \\
\cmark & \xmark & 98.0 & \textbf{98.8} & \textbf{97.4} & 90.0 & \textbf{96.1} \\
\xmark & \cmark & 98.4 & \textbf{98.8} & 96.8 & 89.4 & 95.9 \\
\rowcolor{gray!15}
\cmark & \cmark & \textbf{98.8} & 98.5 & 96.6 & \textbf{90.4} & \textbf{96.1} \\
				
\bottomrule
\end{tabular}%
\end{center}}
\end{minipage}
}
\hspace{2em}
\subfloat[The context grid size $M$. $M{=}2$ preserves finer context using fewer tokens, while $M{=}0$ discards all pruned patches.
\label{tab:grid}
]{
\begin{minipage}{0.4\textwidth}{\begin{center}
\begin{tabular}{c cccc c}
\toprule
\multirow{2}{*}{$M \times  M$}
& \multicolumn{4}{c}{LIBERO Success (\%)}
& \multirow{2}{*}{Acc.(\%)$\uparrow$} \\
\cmidrule(lr){2-5}
& Spatial & Object & Goal & Long & \\
\midrule
$0\times0$          & 98.2 & 96.8 & 96.6 & 88.2 & 95.0 \\
$1\times1$  & 97.6 & \textbf{99.0} & 95.8 & 89.4 & 95.5 \\
\rowcolor{gray!15}
$2\times2$  & 98.0 & 98.8 & \textbf{97.4} & \textbf{90.0} & \textbf{96.1} \\
$4\times4$  & 97.6 & 98.6 & 97.0 & 86.8 & 95.0 \\
$8\times8$  & \textbf{98.8} & 98.2 & 95.4 & 80.0 & 93.1 \\
\bottomrule
\end{tabular}%
\end{center}}
\end{minipage}
}
\vspace{-2em}
\end{table*}

\vspace{-3pt}
\subsection{Ablation Study}\vspace{-3pt}
To analyze the design choices of \ours, we conduct ablation studies at a 25\% token retention ratio on LIBERO. 

\begin{figure}
    \centering
    \includegraphics[width=.85\linewidth]{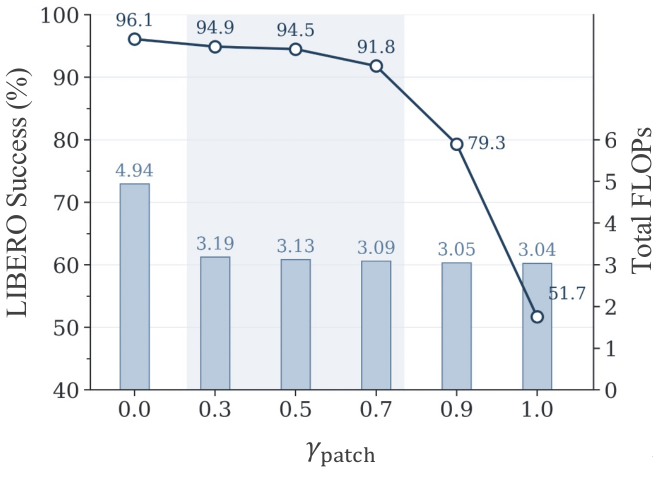}
    \vspace{-.5em}
    \caption{The effectiveness of patch-level retained ratio $\gamma_{\text{patch}}$ with $\pi_{0.5}$ on LIBERO at full retained ratio $25\%$.}
    \label{fig:preratio}\vspace{-3pt}
\end{figure}

\paragrapht{Design Choices.}\vspace{-3pt}
We evaluate whether pruning both views is necessary, since each alone already removes part of the redundancy.
\cref{tab:ablation_choice} shows that pruning a single view retains higher accuracy, but leaves the other untouched and therefore limits the FLOPs reduction.
Pruning both lowers FLOPs to 3.063 at a small accuracy cost, as each view holds redundancy the other cannot reach.
\cref{fig:preratio} then varies how much of the budget is spent before the encoder.
Larger ratios lower encoder computation but rely entirely on action geometry, and accuracy collapses once the post-encoder stage is removed.
We use 70\%, which balances the two.

\paragrapht{Trajectory Horizon.}
We study how much temporal context is needed to identify action-relevant regions in the fixed global view.
Executed poses capture where the manipulator has acted, while extrapolated poses anticipate where it is heading.
In \cref{tab:horizon}, combining executed and extrapolated poses with $(3,3)$ achieves the best performance of 91.8\%, covering both past and upcoming interaction regions.

\paragrapht{Gripper-aware Expansion.}
A projected trajectory specifies points rather than the spatial extent of the interaction. 
We therefore expand each point into a $\kappa\times\kappa$ neighborhood whose size follows the gripper state, since a closed gripper carries an object into view.
As shown in Table~\ref{tab:gripper}, a fixed extent under-covers grasped objects, whereas excessive expansion retains irrelevant background.
We use $\kappa=5$ and $7$ for open and closed gripper states.

\paragrapht{Context Tokens on DINOv2 and SigLIP.}
Context tokens carry a coarse spatial layout, so they are informative only to an encoder that represents spatial structure.
Since OpenVLA-OFT employs two vision encoders, we examine which encoder receives the context tokens.
\cref{tab:merging} shows that merging within DINOv2~\cite{dino} alone achieves 96.1\%, with no additional gain from including SigLIP~\cite{siglip}.

\paragrapht{Grid Cell Size of Context Tokens.}
We study the grid cell size $M$, which divides the whole image into a fixed grid to produce context tokens.
% how much spatial context should be preserved from the pruned regions.
Since context tokens share the patch-level budget with retained patches, a grid allocates more tokens to regions already identified as redundant.
\cref{tab:grid} shows that increasing the grid resolution degrades performance, supporting a coarse $2\times2$ grid.

\begin{table}[t]
\centering
\small
\setlength{\tabcolsep}{4pt}
\caption{
{Performance of our method on $\pi_{0.5}$ for real-robot tasks} at 50\%, 25\%, and 12.5\% visual token retention.
}\vspace{-3pt}
\label{tab:real_results}
% \resizebox{\linewidth}{!}{
\begin{tabular}{l cccc cc}
\toprule
\multirow{2}[3]{*}{Method}
& \multicolumn{4}{c}{Success (\%)$\uparrow$}
& \multirow{2}[3]{*}{\makecell{FLOPs\\(T)$\downarrow$}}
& \multirow{2}[3]{*}{\makecell{Latency\\(ms)$\downarrow$}}\\
\cmidrule(lr){2-5}
& Tool & Doll & Towel & Banana & &  \\
\midrule

% \rowcolor{gray!15}
% \multicolumn{7}{c}{\textit{Upper Bound (100\%)}} \\

$\pi_{0.5}$ (100\%)
& 70 & 60 & 60 & 50 & 5.021 & 174.8  \\

% \rowcolor{gray!15}
\midrule
\multicolumn{7}{l}{\textit{Retained 50\% (50\%$\downarrow$)}} \\

\hspace{.5em}+FastV
& \bf{60} & 50 & \bf{50} & {30} & 3.996 & \underline{167.6}  \\

\hspace{.5em}+SparseVLM
& \bf{60} & 50 & 40 & \textbf{40} & 4.187 & 169.3  \\

\hspace{.5em}+DivPrune
& \bf{60} & 40 & 40 & 20 & \underline{3.869} & 172.3 \\

\hspace{.5em}+VLA-Cache
& 50 & \textbf{60} & \bf{50} & {30} & 4.036 & 169.7 \\

\hspace{.5em}+VLA-Pruner
& 50 & \underline{50} & \bf{50} & 20 & 3.899 & 169.8  \\

\hspace{.5em}+\textbf{\ours}
& \bf{60} & \textbf{60} & \bf{50} & \underline{30} & \textbf{3.710} & \textbf{166.4}  \\

\midrule
\multicolumn{7}{l}{\textit{Retained 25\% (75\%$\downarrow$)}} \\

\hspace{.5em}+FastV
& \textbf{50} & \underline{40} & \underline{30} & 10 & \underline{3.490} & \underline{163.4}  \\

\hspace{.5em}+SparseVLM
& 30 & \underline{40} & \underline{30} & \textbf{20} & 3.675 & 164.5  \\

\hspace{.5em}+DivPrune
& \underline{40} & 30 & 20 & 10 & 3.299 & 163.7  \\

\hspace{.5em}+VLA-Cache
& \underline{40} & \textbf{50} & \textbf{40} & \textbf{20} & 3.799 & 168.8 \\

\hspace{.5em}+VLA-Pruner
& \underline{40} & \underline{40} & \underline{30} & 10 & 3.400 & 167.5  \\

\hspace{.5em}+\textbf{\ours}
& \textbf{50} & \textbf{50} & \textbf{40} & 10 & \textbf{3.066} & \textbf{163.2} \\

\midrule
\multicolumn{7}{l}{\textit{Retained 12.5\% (87.5\%$\downarrow$)}} \\

\hspace{.5em}+FastV
& \underline{10} & \textbf{20} & \textbf{10} & \textbf{10} & 3.239 & \underline{159.1} \\

\hspace{.5em}+SparseVLM
& \underline{10} & \underline{10} & 0 & \textbf{10} & 3.538 & 160.5 \\

\hspace{.5em}+DivPrune
& \textbf{20} & \underline{10} & 0 & 0 & \underline{3.017} & 159.6 \\

\hspace{.5em}+VLA-Cache
& \underline{10} & \textbf{20} & \textbf{10} & \textbf{10} & 3.676 & 162.0 \\

\hspace{.5em}+VLA-Pruner
& \underline{10} & \underline{10} & \textbf{10} & 0 & 3.117 & 161.4 \\

\hspace{.5em}+\textbf{\ours}
& \textbf{20} & \textbf{20} & \textbf{10} & \textbf{10} & \textbf{2.744} & \textbf{158.3} \\

\bottomrule
\end{tabular}
\vspace{-.5em}
\end{table}
\subsection{Real Robot Experiments}
% table 9 
To validate the practical effectiveness of \ours, we further evaluate it on four real-world manipulation tasks using $\pi_{0.5}$. In \cref{tab:real_results}, \ours achieves an average success rate of 50\% at 50\% token retention, compared with 60\% for the vanilla policy, while reducing FLOPs from 5.021 to 3.710. 
At more aggressive retention ratios, \ours maintains average success rates of 37.5\% and 15.0\% at 25\% and 12.5\% retention, respectively, showing competitive or superior performance to the compared pruning methods. 

% latency
% These results demonstrate that \ours preserves real-world manipulation performance under substantial visual token reduction while providing consistent computational savings in physical deployment. 

\vspace{-3pt}
\subsection{Discussion}
% openvla-oft 대비 pi_0.5 에서 speed up이 적은 이유?
% libero 대비 vlabench 성능 떨어지는 이유?
% agressive pruning에서 성능 급격히 떨어지는 이유? 
% action-derived pruning (ours) limitation? 
While \ours consistently reduces computation across VLA models and benchmarks, its practical gains depend on where visual redundancy lies and how much can be safely removed.
On $\pi_{0.5}$, a 38\% FLOPs reduction yields only a 1.03$\times$ speedup, compared with 1.61$\times$ on OpenVLA-OFT, as its action expert integrates a flow field over multiple steps at a cost independent of the visual token count.
The achievable redundancy depends on the camera configuration.
When the gripper occupies a large portion of the wrist view as in VLABench, its patches move with the camera and are conservatively retained.
In addition, extreme pruning ratios gradually trade task performance for efficiency.
% How early pruning is performed introduces a further trade-off.
A patch removed before the encoder significantly reduces visual computation but neither receives global context nor can be recovered downstream.
This motivates our two-stage design: patch-level pruning before the encoder reduces visual computation, while token-level refinement based on encoder attention provides a more informed pruning decision.
Finally, since \ours assumes calibrated cameras and depth sensing, extending the framework to uncalibrated or RGB-only settings is a promising direction for future work.

% \vspace{-.3em}
\section{Conclusion}
% \vspace{-3pt}
We presented \ours, which leverages the previously predicted action chunk as a physically grounded prior available before the current observation is processed to derive view-specific signals:
trajectory prediction with semantic anchors for the third-person view, and motion-compensated redundancy for the wrist view.
By compressing a substantial portion of patches before the vision encoder, \ours reduces the computation of both the visual encoder and the LLM backbone.
Experiments on LIBERO, VLABench, and a real 7-DoF Franka robot show that \ours achieves SOTA efficiency with negligible loss in task success.
% \vspace{-3pt}



% \section*{ACKNOWLEDGMENT}

% The preferred spelling of the word ÒacknowledgmentÓ in America is without an ÒeÓ after the ÒgÓ. Avoid the stilted expression, ÒOne of us (R. B. G.) thanks . . .Ó  Instead, try ÒR. B. G. thanksÓ. Put sponsor acknowledgments in the unnumbered footnote on the first page.

%%%%%%%%%%%%%%%%%%%%%%%%%%%%%%%%%%%%%%%%%%%%%%%%%%%%%%%%%%%%%%%%%%%%%%%%%%%%%%%%

\bibliographystyle{IEEEtran}
\bibliography{IEEEbcpat,mybib}

\end{document}